\documentclass[letterpaper]{article} 
\usepackage{aaai25}  
\usepackage{times}  
\usepackage{helvet}  
\usepackage{courier}  
\usepackage[hyphens]{url}  
\usepackage{graphicx} 
\usepackage{natbib}  
\usepackage{caption} 
\usepackage{algorithm}
\usepackage{algorithmic}
\usepackage{lipsum}
\usepackage{makecell}
\usepackage{multirow}
\usepackage{booktabs}
\usepackage{colortbl}
\usepackage{arydshln}
\usepackage{bm}
\usepackage{xcolor}
\usepackage{graphicx}
\usepackage{subcaption}
\usepackage{tikz}
\usepackage{rotating} 
\usepackage[fixed]{fontawesome5}
\usepackage{nicematrix}
\usepackage{amsmath}
\usepackage{amssymb}
\usepackage{adjustbox}
\usepackage{placeins} 
\usepackage[hidelinks]{hyperref}
\usepackage{newfloat}
\usepackage{listings}
\DeclareCaptionStyle{ruled}{labelfont=normalfont,labelsep=colon,strut=off} 
\floatstyle{ruled}
\newfloat{listing}{tb}{lst}{}
\floatname{listing}{Listing}

\title{Quantum-inspired Interpretable Deep Learning Architecture for Text Sentiment Analysis}
\author{
    Bingyu Li \textsuperscript{\rm 1, \rm 2}\equalcontrib\thanks{Corresponding author},
    Da Zhang\textsuperscript{\rm 2, \rm 3}\equalcontrib,
    Zhiyuan Zhao\textsuperscript{\rm 2}, 
    Junyu Gao\textsuperscript{\rm 2, \rm 3}, 
    Yuan Yuan\textsuperscript{\rm 3}\\
}
\affiliations{
    \textsuperscript{\rm 1}Department of Electronic Engineering and Information Science, University of Science and Technology of China, China 
    \textsuperscript{\rm 2}Institute of Artificial Intelligence (TeleAI), China\\
    \textsuperscript{\rm 3}School of Artificial Intelligence, OPtics and ElectroNics (iOPEN), Northwestern Polytechnical University, China\\
    \texttt{libingyu0205@mail.ustc.edu.cn, dazhang@mail.nwpu.edu.cn, tuzixini@163.com, gjy3035@gmail.com, y.yuan1.ieee@gmail.com}
}

\begin{document}

\maketitle

\begin{abstract}
Text has become the predominant form of communication on social media, embedding a wealth of emotional nuances. Consequently, the extraction of emotional information from text is of paramount importance. Despite previous research making some progress, existing text sentiment analysis models still face challenges in integrating diverse semantic information and lack interpretability.
To address these issues, we propose a quantum-inspired deep learning architecture that combines fundamental principles of quantum mechanics (QM principles) with deep learning models for text sentiment analysis. 
Specifically, we analyze the commonalities between text representation and QM principles to design a quantum-inspired text representation method and further develop a quantum-inspired text embedding layer.
Additionally, we design a feature extraction layer based on long short-term memory (LSTM) networks and self-attention mechanisms (SAMs). 
Finally, we calculate the text density matrix using the quantum complex numbers principle and apply 2D-convolution neural networks (CNNs) for feature condensation and dimensionality reduction.
Through a series of visualization, comparative, and ablation experiments, we demonstrate that our model not only shows significant advantages in accuracy and efficiency compared to previous related models but also achieves a certain level of interpretability by integrating QM principles. Our code is available at \color{blue}{\href{https://github.com/LiBingyu01/QITSA-Quantum-inspired-Interpretable-Deep-Learning-Architecture-for-Text-Sentiment-Analysis}{QISA}}.
\end{abstract}

\section{Introduction}
\quad
The volume of textual information generated online is continuously increasing \cite{xu2019chinese}. Analyzing the sentiment of this textual information is essential for understanding social opinion trends and overall product evaluations \cite{niu2021review}. Consequently, sentiment analysis technology is becoming increasingly vital in fields such as social media analysis, market sentiment tracking, and public opinion monitoring \cite{kim2014convolutional}. Text sentiment analysis, a method for extracting emotional tendencies from texts, has garnered widespread attention \cite{wang2017combining}.

\begin{figure}
\centering
\includegraphics[width=\linewidth]{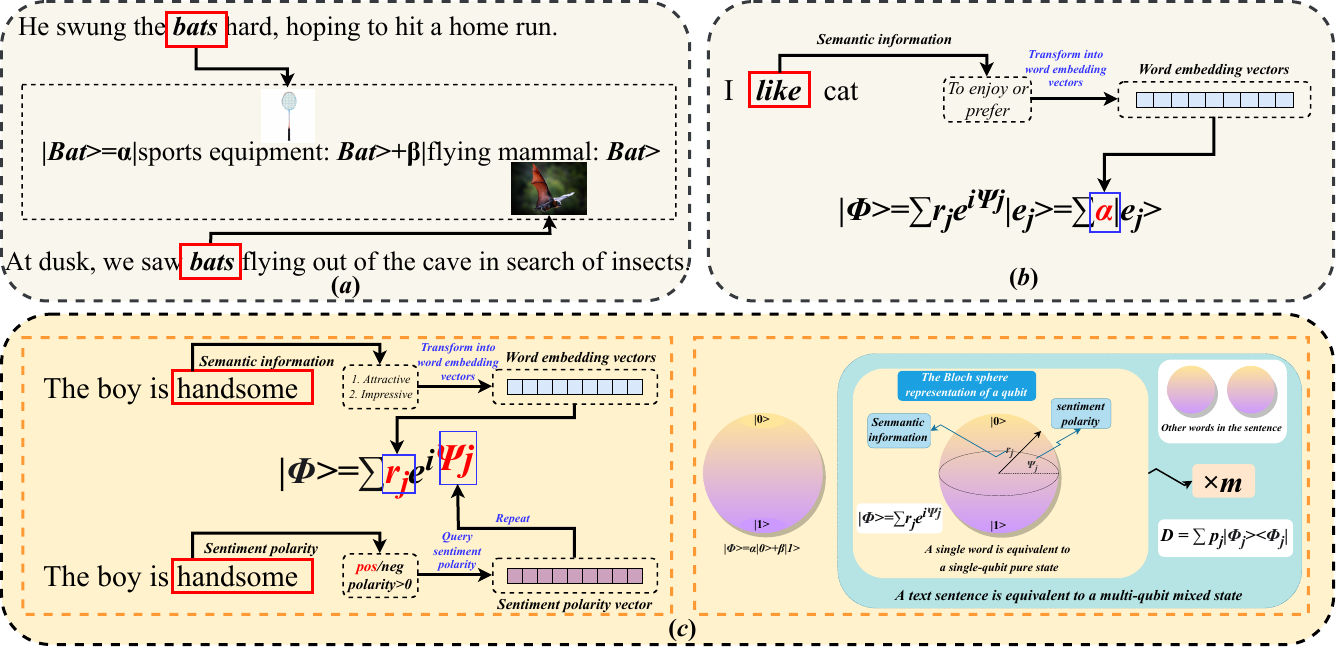}
\caption{Illustration of Quantum-Inspired Language Representation. (a) Synonymy corresponding to quantum superposition principle. (b) Embedding semantic information into complex numbers (ignoring multimodal information). (c) Left: Embedding both semantic and sentiment information. Right: Bloch Sphere Representation for the QITSA.}
\label{fig:fig_1}
\end{figure}

Mainstream text sentiment analysis models follow three primary routes. The first approach is based on sentiment lexicons, which rely on manually annotated word sentiment lexicons \cite{rice2021corpus}. Although sentiment lexicons have made significant progress, the precision of this method depends entirely on the coverage and accuracy of the sentiment lexicon, making it overly reliant on its construction.
To enhance model generalizability, classic machine learning methods have been explored. However, these methods often ignore the contextual semantic dependencies of the text \cite{wang2017combining, muthusankar2023bidrn}. To address this issue, many researchers have conducted in-depth studies using deep learning models, achieving considerable success with models based on CNNs, Recurrent Neural Networks (RNNs), and SAMs. These models have demonstrated excellent performance and have been widely
applied.

Despite the contributions of previous models to improving analysis accuracy, two notable shortcomings remain: 1) prior models fail to integrate multiple semantic information, and 2) the models lack interpretability. Text sentiment analysis tasks need to consider not only words' semantic dependency but also sentiment-bearing words' dependency \cite{hu2004mining, sachin2020sentiment}. Therefore, a comprehensive analysis of various semantic features is necessary. However, existing methods struggle to integrate multiple features for comprehensive analysis \cite{shobana2021efficient, huang2021attention}. Additionally, the underlying mechanisms of these models remain difficult to interpret, and the design motivations lack deep physical underpinnings.

In this paper, we first analyze the commonalities between text representation and QM principles from the perspective of combining text features with QM principles (Fig.\ref{fig:fig_1}). In light of this analysis, we propose a text sentiment analysis architecture, dubbed \textbf{Q}uantum-inspired \textbf{I}nterpretable \textbf{T}ext \textbf{S}entiment \textbf{A}nalysis Architecture (QITSA).
Drawing on the commonalities between text representation and QM principles, we develop a quantum-inspired text representation method and a corresponding word embedding layer. Unlike previous forms that embedded only one type of information (Fig.\ref{fig:fig_1}(b)), This layer leverages complex numbers to integrate various semantic information, achieving adaptive fusion and embedding multiple types of information (Fig.\ref{fig:fig_1} (c)). 
To further extract textual features, we designed a text feature extraction layer based on LSTM networks and SAMs, which model the text's contextual and global dependencies, thereby enabling deeper feature extraction.
We achieve the final results by representing text vectors as real and imaginary parts in two separate processing tracks, calculating density matrices for each, and using two-dimensional convolutional neural networks (2D-CNNs) for final feature fusion.
To validate the effectiveness of the proposed model, we conducted tests on multiple text sentiment analysis datasets, demonstrating that our framework outperforms related algorithms in the field. To further assess the effectiveness of modal fusion and the contribution of the proposed modules, we conducted visualization experiments and ablation studies, exploring the impact of various factors on model performance.

In summary, our contributions can be summarized as follows:
\begin{itemize}
    \item We analyze the commonalities between fundamental QM principles and text representation, proposing a quantum-inspired deep learning architecture that enhances model accuracy and interpretability.
    \item A quantum-inspired text sentiment word embedding layer is designed, integrating semantic information and sentiment polarity information using quantum complex principles to guide sentiment analysis tasks.
    \item We develop a quantum-inspired feature extraction layer based on the quantum complex-number principle and deep learning model for feature concentration and classification.
    \item Extensive visualization of text vectors and numerical comparison analysis verify the model's effectiveness.
\end{itemize}
\section{Related Work}
\subsection{Text Sentiment Analysis}
\quad
Text Sentiment analysis has been studied using various methods, including lexicons-based and deep-learning techniques. 
Lexicons-based methods which rely on predefined lexicons were among the earliest to be developed. 
Based on this approach, some studies have applied it to domain-specific semantic analysis \cite{rice2021corpus} and customer reviews summary, while others have extended it to different languages \cite{xu2019chinese}. However, the lexicons-based method faces challenges in scalability and in effectively addressing semantic dependencies and ambiguities in text.
To tackle these issues, more deep-learning methods have been thoroughly investigated. CNNs, known for their ability to extract both local and global features, have been applied to text sentiment analysis \cite{kim2014convolutional, wang2017combining}. To capture longer semantic dependencies, RNNs have been developed, with Long Short-Term Memory (LSTM) networks showing excellent performance in establishing long-term semantic dependencies \cite{muthusankar2023bidrn, shobana2021efficient}. However, RNN-based methods often suffer from gradient vanishing problems, making it difficult to model dependencies in long text sequences.
To address this issue, attention mechanisms have been introduced to model the global dependencies of the text \cite{vaswani2017attention, huang2021attention}.

Despite the significant effectiveness of the aforementioned methods, they still struggle to integrate multiple sentiment information, and the models lack interpretability.
In this article, we design QITSA based on the QM principles to integrate various types of semantic information and enhance the interpretability of deep learning models.
\subsection{Quantum-Inspired Deep Learning Model}
\quad
The QM principles are increasingly applied to natural language processing (NLP) tasks due to their commonalities with text representation. Initially, pure quantum computing algorithms were extended to text sentiment analysis \cite{zhang2018end, li2019cnm}. Subsequently, researchers developed Lambeq \cite{kartsaklis2021lambeq}, an advanced Python library designed to facilitate the implementation of quantum NLP models.
While pure quantum computing algorithms enhance model interpretability, they face limitations in processing large-scale textual data. Therefore, researchers have proposed quantum-inspired deep learning models by combining the interpretability of quantum mechanics with the ability of deep learning to handle large-scale data and benefit from hardware acceleration \cite{liu2023quantum, santur2019sentiment}.
Leveraging these characteristics, some studies have introduced quantum-inspired complex word embeddings \cite{li2018quantum}, which represent words in a complex vector space, capturing intricate semantic relationships more effectively than traditional embeddings. In the field of text sentiment analysis, researchers have designed two end-to-end quantum-inspired deep neural networks for text classification, illustrating how the QM principle can be integrated into neural network architectures to improve performance \cite{shi2021two}. Additionally, other researchers have been inspired by these commonalities to design more fine-grained quantum-inspired text sentiment analysis models \cite{wang2023quantum}.

In this paper, we explore the enhancement of multimodal sentiment semantic information embedding and fusion through the QM principles. Additionally, we have designed an end-to-end quantum-inspired deep learning architecture that combines the interpretability of quantum computing with the superior feature extraction capabilities of deep learning.
\section{Method}
\subsection{Preliminary}
\quad
\textbf{Quantum Superposition.} The quantum superposition principle is one of the foundations of quantum mechanics, describing the superposition of states that a microscopic particle can be in. According to the superposition principle, a quantum system can exist in a linear combination of multiple states until it is observed or measured.

An individual quantum is typically represented as:
\begin{equation}
|\psi\rangle = \sum_{i} c_i |i\rangle, \quad
\end{equation}
where $c_i$ represents the probability amplitudes for the quantum to be in state $|i\rangle$ when measured.

Generally, a quantum physical system often contains multiple particles. We refer to this system as a mixed state, while a single-quantum system is referred to as a pure state. To describe a quantum system composed of multiple particles, the superposition state of a quantum can be described as follows:  
\begin{equation}
|\psi\rangle = \sum_j r_j e^{i \beta_j} |\psi_j\rangle,
\label{eq:eq_2}
\end{equation}  
where $i$ is the imaginary unit, $r_j$ and $\beta_j$ represent the amplitude and phase of the $j$-th orthogonal superposition state, respectively, and $|\psi_j\rangle$ represents the orthonormal basis state in the Hilbert space.  

\textbf{Quantum Density Matrix.} The quantum density matrix is a mathematical tool used to describe the state of a quantum system. A quantum physical system containing multiple particles can be represented by a density matrix, which is expressed as:
\begin{equation}
\rho = \sum_{i} p_i |\psi_i\rangle\langle\psi_i|, \quad
\end{equation}
where $p_i$ represents the probability value, satisfying $\sum_{i} p_i = 1$, $|\psi_i\rangle$ is the superposition state of a single quantum in Eq. \ref{eq:eq_2}, and $\langle\psi_i|$ is the transpose of $|\psi_i\rangle$. $|\psi_i\rangle$ is a pure state vector with probability $p_i$. The density matrix $\rho$ is symmetric, semi-positive definite, and has a trace of 1, i.e., $\text{Tr}(\rho) = 1$. According to the Gleason theorem, there is a bijective correspondence between quantum probability measures $P$ and density matrices $\rho$, denoted as $P \leftrightarrow \rho$.

In text representation, we consider a sentence as a multi-body system, with each word corresponding to a component in the system. By using the quantum density matrix, we can more accurately describe the relationships and interactions between words in a sentence.
\begin{figure}
\centering
\includegraphics[width=0.95\linewidth]{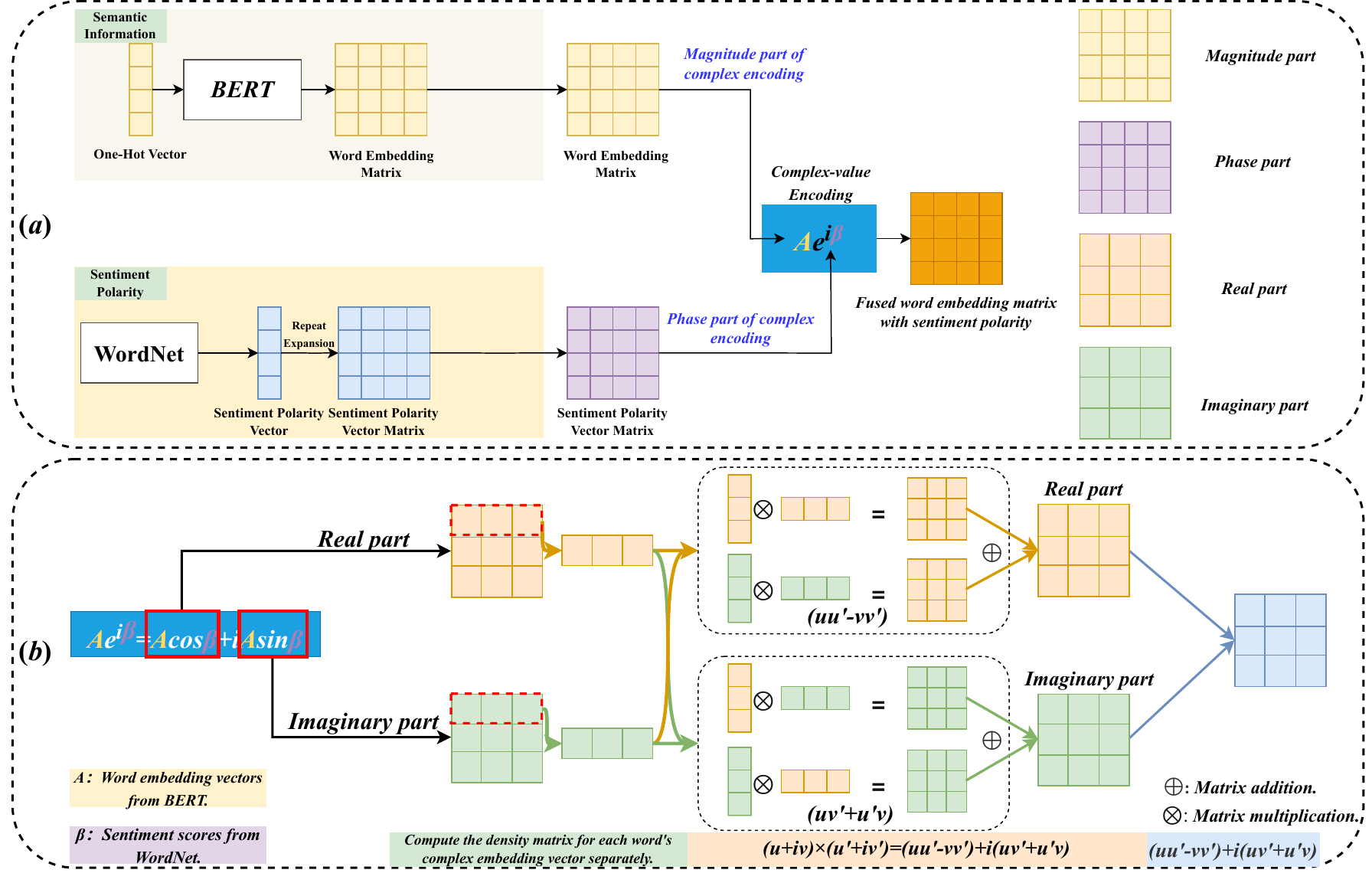}
\caption{The QM principle word embedding layer. (a) The principle of quantum complex numbers integrating multiple semantic information, and (b) Euler's formula for calculating the imaginary and real parts of the semantic vector matrix.}
\label{fig:fig_2}
\end{figure}
\begin{figure*}
\centering
\includegraphics[width=\linewidth]{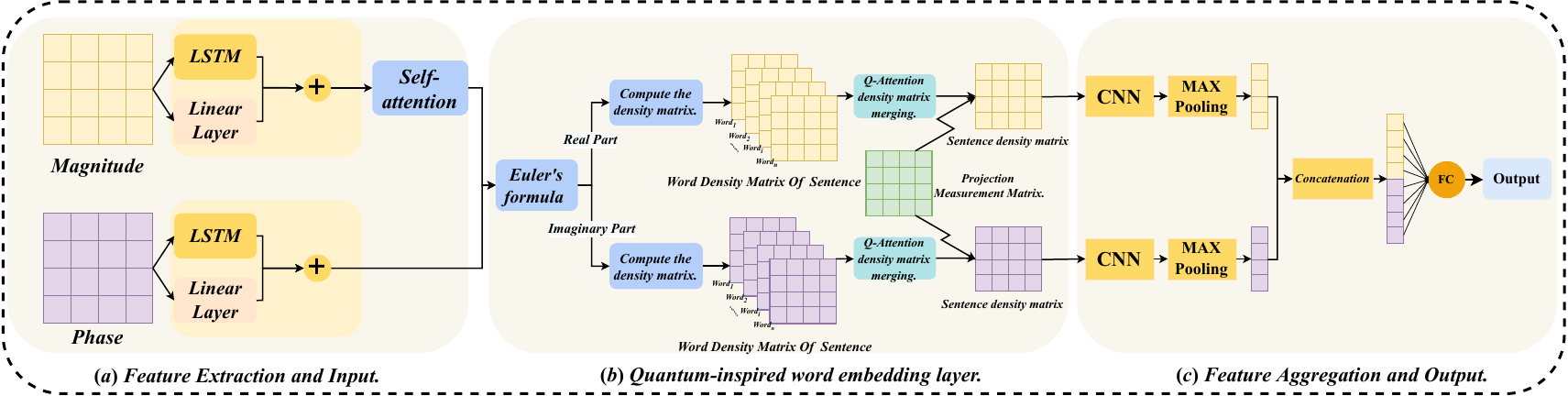}
\caption{Overall process framework and detailed structure diagram. (a)(b)(c) jointly form the common framework of the article.}
\label{fig:fig_3}
\end{figure*}
\subsection{Quantum-Inspired Language Representation}
In sentiment analysis tasks, accurately understanding the meaning of words is crucial. Quantum mechanics offers a unique perspective by viewing the multiple meanings of a word as a quantum superposition. We represent the semantic state of a sentence using a density matrix, capturing word associations and interactions.

The multiple meanings of a word can be viewed as a quantum superposition (Fig.\ref{fig:fig_1}(a) and (b)), with the word's embedding vector $|\mathbf{v}_j\rangle$ mapped to a Hilbert space, forming basic events used to construct a density matrix $\mathbf{\rho}_j = |\mathbf{v}_j\rangle \langle\mathbf{v}_j|$. Multiple words in a sentence form a quantum mixed system, with the sentence's density matrix being the weighted sum of individual word density matrices:
\begin{equation}
\mathbf{\rho} = \sum_j w_j \mathbf{\rho}_j,
\end{equation}
where $w_i$ is the weight of word $i$ in the sentence.

As demonstrated in Fig.\ref{fig:fig_1}(c), the emotional polarity scores of words are regarded as phase information, while semantic information serves as amplitude. By calculating these, we obtain a quantum superposition text representation that integrates both semantic and emotional information:
\begin{equation}  
\begin{aligned}  
\mathbf{v}_j &= \textit{SemInfo}(\mathbf{v}_j) \cdot e^{i \cdot \textit{EmoPhase}(\mathbf{v}_j)} = r_j \cdot e^{i\cdot\beta_{j}},
\end{aligned}  
\label{eq:eq_5}
\end{equation}  
where, \(SemInfo(\cdot)\) and \(EmoPhase(\cdot)\) obtain the semantic and sentiment information of \(|\mathbf{v}_i\rangle\). This method can capture complex interactions between words, providing a more comprehensive representation of sentence semantics and emotions, and potentially improving model performance and accuracy.

\subsection{Quantum-Inspired Text Embedding Layer}
\quad
Using the quantum exponential representation, where a complex number has an amplitude and a phase, we treat word embedding information as the amplitude and sentiment information as the phase, as shown in Fig.\ref{fig:fig_2} (a). This fusion into a complex-valued word vector is expressed as Eq. \ref{eq:eq_5}. 

For semantic information, we extract the frequency of each word in the text to obtain its one-hot vector representation which is input into the BERT model \cite{devlin2018bert} to obtain the word embedding vector matrix. For sentiment information, we extract the polarity scores of multiple words from large lexicons (such as WordNet) \cite{rice2021corpus,husnain2021systematic,alrashidi2020reflective}.

Directly handling the complex representation of words on classical computers may pose computational complexity challenges. Therefore, we adopt Euler's formula, $Ae^{i\beta} = A(\cos\beta + i\sin\beta)$, to convert the complex exponential form into its real and imaginary components representation. Specifically, for a given word embedding vector, we compute its real part $\text{Re}(|\mathbf{v}_j\rangle)$ and imaginary part $\text{Im}(|\mathbf{v}_j\rangle)$. This transformation simplifies the originally complex exponential operations into more tractable real and imaginary parts.  
\begin{equation}  
y = \sum_j \text{Re}(|\mathbf{v}_j\rangle) + i\text{Im}(|\mathbf{v}_j\rangle) 
\end{equation}
Subsequently, we map the word representation vectors to quantum events and calculate the quantum density matrix as follows.
\begin{equation}
\begin{aligned}  
\rho  &= \sum_{j}  |\mathbf{v}_j\rangle\langle\mathbf{v}_j|  \\  
  &= \sum_{j} | r_j (\cos(\beta_j) + i\sin(\beta_j))\rangle\langle r_{j} (\cos(\beta_{j}) + i\sin(\beta_{j}))| \\  
  &= \sum_{j} \left[ r_j r_{j} \left( |\cos(\beta_j)\rangle\langle\cos(\beta_{j})| - |\sin(\beta_j)\rangle\langle\sin(\beta_{j})| \right) \right. \\  
  &\quad \left. + i [r_j r_{j} \left(|\sin(\beta_j)\rangle\langle\cos(\beta_{j})| + |\cos(\beta_j)\rangle\langle\sin(\beta_{j})| \right) \right] \\  
  &= \sum_{j} [r_j r_{j}\left( \text{Re}(\mathbf{v}_j) \text{Re}(\mathbf{v}_j) - \text{Im}(\mathbf{v}_j) \text{Im}(\mathbf{v}_j) \right)] \\  
  &\quad + i [r_j r_{j}\left( \text{Re}(\mathbf{v}_j) \text{Im}(\mathbf{v}_j) + \text{Im}(\mathbf{v}_j) \text{Re}(\mathbf{v}_j) \right)] \\  
  &= \sum_{j} r_j r_{j} \left( \rho_j^{\text{real}} + i\rho_j^{\text{imag}}\right),  
\end{aligned}
\end{equation}

We leverage the values of the real and imaginary parts to perform the corresponding operations and obtain the density matrix, the calculation process is shown in Fig.\ref{fig:fig_2} (b). 
Although the final complex density matrix is the sum of the real and imaginary parts, extracting information from the real and imaginary parts separately and then fusing them can capture more information. The practical implementation is:
\begin{equation}
\rho = \sum_i w_i (\rho_i^{\text{real}} + \rho_i^{\text{imag}}),
\end{equation}
where \(\rho_i^{\text{real}}\) and \(\rho_i^{\text{imag}}\) are the density matrices of the real and imaginary parts, and \(w_i\) are learnable parameter values.

\subsection{Quantum-Inspired Feature Extraction Layer for Text Sentiment Analysis}
\quad
Based on the aforementioned architecture, we design an end-to-end quantum-inspired deep learning model for text sentiment analysis (Fig. \ref{fig:fig_3}(a)).

\subsubsection{Feature Extraction and Input.} As shown in Fig.\ref{fig:fig_3}(a), before applying quantum-inspired word embedding to the word vector matrix, we utilize an LSTM-attention mechanism to model the dependencies within the text \cite{shobana2021efficient}.

For any given semantic and sentiment information of a word \(|v_j\rangle\), we establish the context dependency:
\begin{equation}
\centering
\begin{aligned}
r_j = LSTM(SemInfo(|v_j\rangle)) \\
\beta_{j} = LSTM(EmoPhase(|v_j\rangle))
\end{aligned}
\end{equation}
We use LSTM to extract the semantic information of amplitude and the sentiment information of phase, thus avoiding the problem of syntactic confusion caused by random combinations of semantics by other non-sequential models, ensuring that "I like cat" does not become "Like I cat" or "Cat like I," etc.

In addition to LSTM, we introduce the self-attention mechanism to extract the key semantic information of amplitude \(r\). The self-attention mechanism focuses on the keywords in the sentence, which is crucial for judging the sentiment polarity of the sentence. The calculation of the self-attention mechanism is as follows:
\begin{equation}
\centering
\begin{aligned}
E &= rW_qW_k^\intercal, \\
F(r) &= softmax\left(\frac{E}{\sqrt{d}}\right), \\
O(r) &= F(r)W_v,
\end{aligned}
\end{equation}
where \(W_q\), \(W_k\), \(W_v\) are trainable weight matrices, \(d\) is the dimension of the input word embedding vectors, and \(X\) is the input word embedding vector matrix.

\subsubsection{Quantum-inspired word embedding layer.} After obtaining the semantic and sentiment information from LSTM and the self-attention mechanism. we calculate the words density matrix representation using Equ. (7) and fuse them into a sentence text embedding matrix representation. This fusion enhances the model's ability to understand and process the sentiment information in NLP using the QM principle.

For sentence density matrix fusion, we propose a Q-Attention mechanism. This mechanism calculates the attention scores between two density matrices and weights their contributions in the fusion process, thereby enhancing the model's attention to keywords in the text. The calculation of attention scores is as follows:
\begin{equation}
F_Q(\rho_i, \rho_j) = softmax\left(\sum_{k=1}^{n} \text{Tr}(\rho_i \odot \rho_j)\right),
\end{equation}
where, \(\odot\) is Hadamard product, \(\rho_i\) and \(\rho_j\) are the density matrices of two words, $Tr$ is to take the diagonal element.

The output matrix is the weighted sum of the density matrices, with the weights as the attention scores:
\begin{equation}
\rho_{1} = \sum_{i,j} F_Q(\rho_i, \rho_j) \cdot \rho,
\end{equation}
where \(\rho_{1}\) is the output density matrix, \(\rho\) is the density matrices of a sentence, and \(F_Q(\rho_i, \rho_j)\) is the attention score between two words of the sentence.

\subsubsection{Feature Aggregation and Output.} In the final step in Fig.\ref{fig:fig_3}(c), we use CNNs to further extract features from the density matrix representation of the text \cite{behera2021colstm}.

Because the density matrix serves as a two-dimensional representation of text, we use 2-D CNNs to extract information. The calculation of the convolutional layer is essentially a special fully connected layer:
\begin{equation}
\rho_{2} = Conv_{2D}(\rho_{1}),
\end{equation}
After the CNNs obtain the feature map, we use max-pooling to generate the vector representations for imaginary and real parts:
\begin{equation}
f = MaxPooling(\rho_{2}).
\end{equation}

Finally, we concatenate these vectors to form the final text embedding representation, which is then passed through a fully connected layer to obtain the final result.
\begin{equation}
Output = FC(Concatenate(f_{img}, f_{real})).
\end{equation}
The overall architecture of the end-to-end deep learning model for text sentiment analysis is illustrated in Fig.\ref{fig:fig_3}.

\section{Experimental Result}

\subsection{Experimental Details}
Given that all experimental parameters are consistent. The experiments were conducted on a single NVIDIA 3090 GPU. All experiments used AdamW as the optimizer, running for 51 epochs. We set the batch size to 16 and the learning rate to 0.001. The binary cross-entropy loss function was chosen as the loss function.
The trained GloVe word vector model \cite{pennington2014glove} is selected as the pre-trained model, with an output dimension of 100, used as the embedding data for the amplitude part of the embedding layer.

\subsection{Dataset}
Five different binary classification benchmark test sets \cite{shi2021two} are chosen to evaluate the performance of the proposed model. Table \ref{tab:dataset_info} provides specific information about the test sets, including the dataset splits and labels, for comparison with baseline models. The details of the datasets can be found in Appendix A.
\begin{table}[htbp]
\centering
\small
\begin{tabular}{cccccc}
\hline
Dataset & Train & Validation & Test & Total & Labels \\
\hline
MR & 8530 & 1065 & 1067 & 10662 & Pos/Neg \\
SST & 67349 & 872 & 1821 & 70042 & Pos/Neg \\
SUBJ & 8000 & 1000 & 1000 & 10000 & Sub/Obj \\
CR & 3024 & 364 & 384 & 3772 & Pos/Neg \\
MPQA & 8496 & 1035 & 1072 & 10623 & Pos/Neg \\
\hline
\end{tabular}
\caption{Summary of Dataset Information}
\label{tab:dataset_info}
\end{table}

\subsubsection{Evaluation Metric}
For Evaluation, accuracy, recall, and F1 score are chosen as metrics to evaluate the model's generalization ability. They are calculated as follows:
\begin{equation}
    \text{Accuracy} = \frac{TP + TN}{TP + TN + FP + FN},
\end{equation}
\begin{equation}
    \text{Recall} = \frac{TP}{TP + FN},
\end{equation}
\begin{equation}
    \text{F1 Score} = 2 \times \frac{\text{Precision} \times \text{Recall}}{\text{Precision} + \text{Recall}},
\end{equation}
where TP (True Positive) is the number of correctly predicted positive cases, TN (True Negative) is the number of correctly predicted negative cases, FP (False Positive) is the number of incorrectly predicted positive cases, and FN (False Negative) is the number of incorrectly predicted negative cases.

\subsection{Experimental Results}
\subsubsection{Results on different datasets.} 
The comparative analysis of different models (table \ref{tab:com_five_dataset}) on test accuracy across five benchmark datasets reveals that our proposed model consistently outperforms others, achieving the highest accuracy in most datasets. Specifically, our model achieves an accuracy of 80.30 on MR, surpassing the second-best model, CE-Mix, by 0.5. On SST, our model scores 84.76, which is slightly lower than CICWE-QNN's 85.0 but still significantly higher than most other models. In the SUBJ dataset, our model achieves 92.96, closely following CICWE-QNN's 93.2. For the CR dataset, our model attains 85.16, outperforming the next best model, CICWE-QNN, by 1.86. On the MPQA dataset, our model achieves 87.50, which is the highest among all models and surpasses CICWE-QNN's 87.2 by 0.3. 
Our model secures the top final rank. This comprehensive evaluation underscores the superior efficacy of our model in text classification, providing valuable insights for future research in this domain.
\begin{table}[htbp]
\centering
\small
\begin{tabular}{lccccccc}
\hline
Model                & MR & SST & SUBJ & CR & MPQA \\ 
\hline
CaptionRep BOW       & 61.9 & - & 77.4 & 69.3 & 70.8 \\
DictRep BOW          & 76.7 & - & 90.7 & 78.7 & \underline{87.2}\\
Paragram-Phrase      & - & 79.7 & - & - & - \\
CE-Sup               & 78.4 & 82.6 & 92.6 & 80 & 85.7 \\
CE-Mix               & \underline{79.8} & 83.3 & 92.8 & 81.1 & 86.6 \\
ICWE-QNN             & 78.6 & 84.2 & 92.6 & 82.6 & 86.8 \\
CICWE-QNN            & 78.3 & \textbf{85.0} & \textbf{93.2} & \underline{83.3} & \underline{87.2}\\
\hline
Ours   & \textbf{80.3} & \underline{84.76} & \underline{92.96} & \textbf{85.16} & \textbf{87.5}\\ 
\hline
\end{tabular}%
\caption{Comparison of the related models on accuracy across five benchmark datasets.}
\label{tab:com_five_dataset}
\end{table}

\subsubsection{Recall on different datasets.} As shown in table \ref{tab:f1_recall}, our model demonstrates consistent performance across five benchmark test sets (MR, SST, SUBJ, CR, MPQA) with F1 Scores ranging from 72.77 to 91.04 and Recall Rates from 81.24 to 93.26. Its high precision and recall, particularly on the SUBJ dataset (F1 Score: 91.04, Recall: 93.26), highlight its robustness and reliability in diverse text classification tasks. 
\begin{table}[h]
\centering
\small
\begin{tabular}{ccccccc}
\hline
Metric & MR & SST & SUBJ & CR & MPQA\\ \hline
F1 Score & 72.77 & 81.78 & 91.04 & 85.16 & 75.81 \\ 
Recall  & 81.24 & 92.62 & 93.26 & 85.16 & 83.50 \\ \hline
\end{tabular}
\caption{Comparison of Maximum F1 Score and Recall Rate of Models on Five Benchmark Test Sets.}
\label{tab:f1_recall}
\end{table}
\paragraph{Visualization.}

\begin{figure}[h]
\centering
\includegraphics[width=0.92\linewidth]{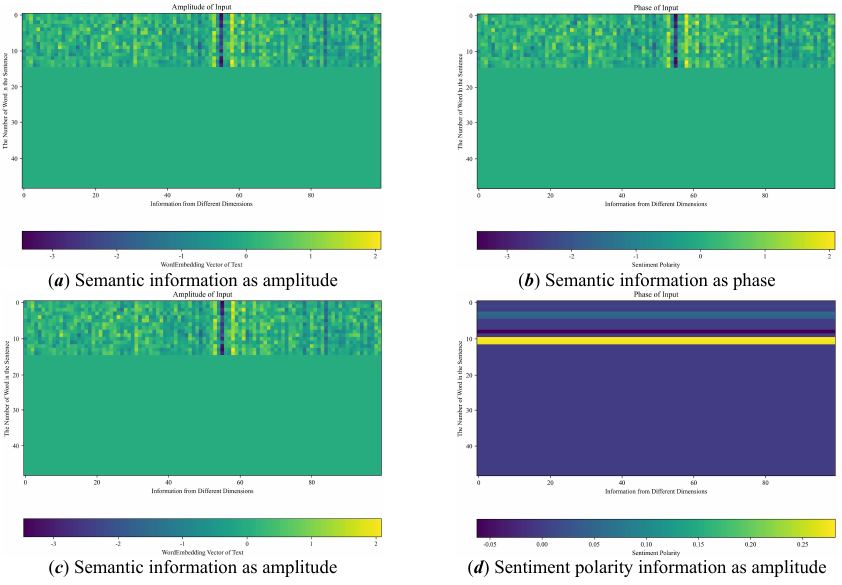}
\caption{Different input combinations. In the first row (Example 1), both amplitude and phase are semantic information. In the second row (Example 2), the amplitude is semantic information, while the phase is sentiment polarity information.}
\label{fig:fig_44}
\end{figure}

To better compare the effectiveness of quantum-inspired text embedding layers that integrate various semantic information for subsequent classification, we input different information combinations as shown in Fig. \ref{fig:fig_44}.

\begin{figure}[ht]
\centering
\includegraphics[width=0.92\linewidth]{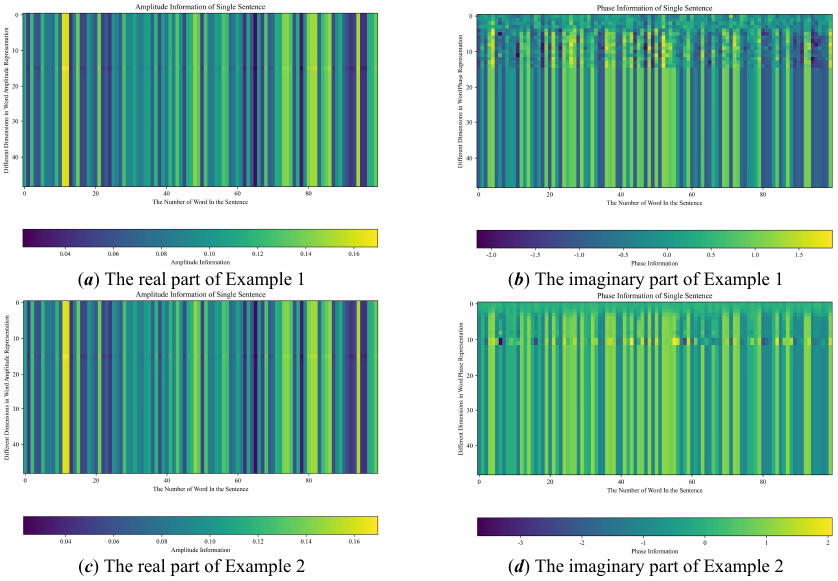}
\caption{Real and imaginary part vectors after LSTM and self-attention processing.}
\label{fig:fig_45}
\end{figure}

After modeling with LSTM and attention mechanisms, the analysis results are displayed in Fig. \ref{fig:fig_45}. The figures show the model's capability to establish global dependency and long-context dependency.

\begin{figure}[ht]
\centering
\includegraphics[width=0.92\linewidth]{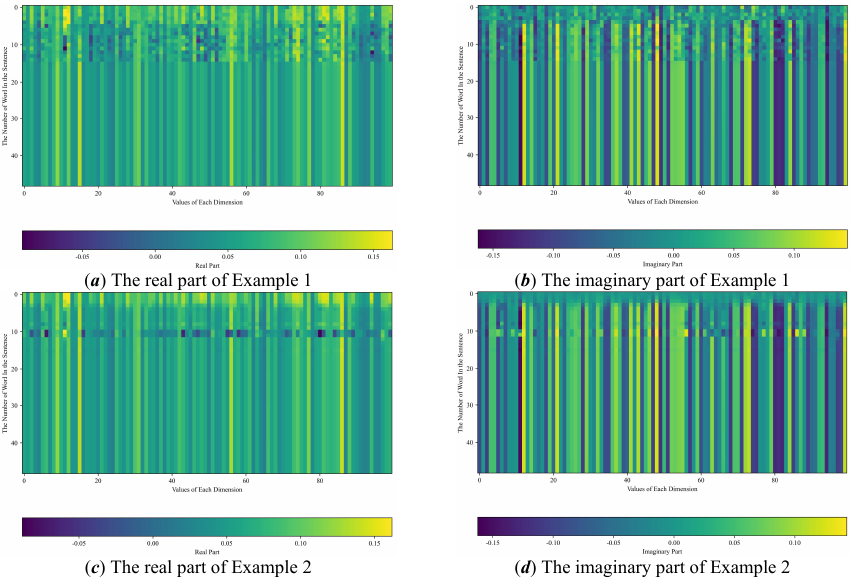}
\caption{Real and Imaginary parts after the quantum-inspired information embedding layer. Example 2 can better highlight areas with significant emotional changes while suppressing areas with weak emotional signals.}
\label{fig:fig_46}
\end{figure}

Furthermore, After processing the two input combinations using the quantum-inspired word embedding method, we visualized the real and imaginary parts of the information (Fig.\ref{fig:fig_46}). Comparing Example 1 and Example 2 reveals that the sentiment integration method, guided by sentiment polarity scores, enables the model to focus on words with significant sentiment polarity changes.
\begin{table*}[htbp]
\centering
\small
\begin{tabular}{lccccccc}
\hline
Input (Amplitude+Phase)                         & MR                                    & SST                                   & SUBJ                                  & CR                                    & MPQA                                  & Avg. Rank                            & Final Rank                              \\ \hline
Word\_Embedding + BERT        & 78.51                                 & 83.85                                 & \textbf{92.96}       & 84.38                                 & \underline{86.94}      & \underline{2.4}        & \underline{2}         \\ 
Word\_Embedding + Word\_Embedding & 74.19                              & 81.83                                 & 91.67                                 & 81.77                                 & 84.14                                 & 5.4                                    & 6                                       \\ 
Word\_Embedding + Sentiment   & 74.13                                 & 83.45                                 & 91.87                                 & 81.77                                 & 84.14                                 & 5.2                                    & 5                                       \\ 
BERT + Word\_Embedding        & \underline{79.07}      & 83.47                                 & 92.56                                 & 83.85                                 & 86.75                                 & 3.4                                    & 4                                       \\
BERT + BERT                   & 78.42                                 & \underline{84.29}      & 92.36                                 & \underline{84.90}      & \underline{86.94}       & 2.8        & 3           \\ 
BERT + Sentiment              & \textbf{80.30}       & \textbf{84.76}       & \textbf{92.96}       & \textbf{85.16}       & \textbf{87.50}       & \textbf{1}            & \textbf{1}            \\ \hline
\end{tabular}
\caption{Ablation experiments on the impact of different embedding data on test accuracy}
\label{tab:diff_embeddata}
\end{table*}
\begin{table*}[htbp]
\centering
\small
\begin{tabular}{lccccccc}
\hline
Model              & MR    & SST   & SUBJ  & CR    & MPQA  & Avg. Rank & Final Rank \\ \hline
CNN-Maxing Pooling & \textbf{80.3} & 84.76 & \textbf{92.96} & \underline{84.90}  & \underline{87.50}  & \underline{1.89}        & \underline{2}         \\ 
Maxing Pooling     & 78.89 & 84.79 & 92.26 & 84.64 & 86.94 & 3.37         & 4          \\ 
CNN-Diagonal       & \underline{80.15} & \underline{85.10}  & \underline{92.66} & \textbf{85.68} & \textbf{87.59} & \textbf{1.52}        & \textbf{1}          \\ 
Diagonal           & 79.77 & \textbf{85.94} & 92.26 & \textbf{85.68} & 86.75 & 2.05         & 3          \\ \hline
\end{tabular}
\caption{Ablation experiments on different dimensionality reduction operations on test accuracy}
\label{tab:dim_red}
\end{table*}
\begin{table}[htbp]
\centering
\begin{tabular}{lccccc}
\hline
Model       & MR    & SST   & SUBJ  & CR    & MPQA  \\ \hline
Q-Attention & \textbf{80.30}  & 84.76 & \textbf{92.96} & \textbf{84.90} & \textbf{87.50} \\ 
Mean        & 78.56 & \textbf{85.91} & 92.76 & 84.64 & 86.57 \\ \hline
\end{tabular}
\caption{Ablation experiments of Q-Attention effectiveness on different datasets}
\label{tab:Q_attn}
\end{table}
\subsection{Ablation Experiment}
This section conducted a series of extended experiments on the proposed model. By comparing the experimental results, we deeply analyzed the specific contributions of each module to the model performance.
\subsubsection{Impact of Different Embedding Data on Test Accuracy.}
To verify the enhancement effect of sentiment embedding on sentiment polarity analysis, we conducted six experiments to study the impact of different embedding data on sentiment polarity analysis (Table \ref{tab:diff_embeddata}). The data included sentiment polarity score vectors (Sentiment), trainabd word embedding vectors (Word\_Embedding), and BERT word embedding representations (BERT), combined in various ways.
Integrating BERT word embeddings with sentiment polarity information consistently improved performance across multiple datasets. For the MR dataset, this combination achieved the highest test accuracy of 80.3\%, demonstrating its effectiveness. Similar enhancements were observed on the CR, SUBJ, MPQA, and SST datasets, indicating that sentiment polarity information complements BERT embeddings well and adapts to various dataset characteristics, ultimately enhancing sentiment analysis model performance.
\subsubsection{Effectiveness Analysis of Q-Attention Mechanism.} 
Q-Attention demonstrated effectiveness in sentiment text analysis across various datasets. It notably improved accuracy on MR, SUBJ, and MPQA datasets, showcasing its ability to capture sentiment information, especially with keywords. However, its performance on the SST dataset was comparatively lower, suggesting the need to choose the appropriate attention mechanism based on dataset characteristics.
\subsubsection{Analysis of the Effects of Different Feature Concentration and Extraction Modules.} 
As shown in table \ref{tab:dim_red}, there are four different types of modules, including two-dimensional convolutional max-pooling operation (CNN-Maxing Pooling), max-pooling operation (Maxing Pooling), diagonal data acquisition operation with two-dimensional convolution (CNN-Getting Diagonal Data), and diagonal data acquisition operation (Getting Diagonal Data).
\begin{figure}
\centering
\includegraphics[width=\linewidth]{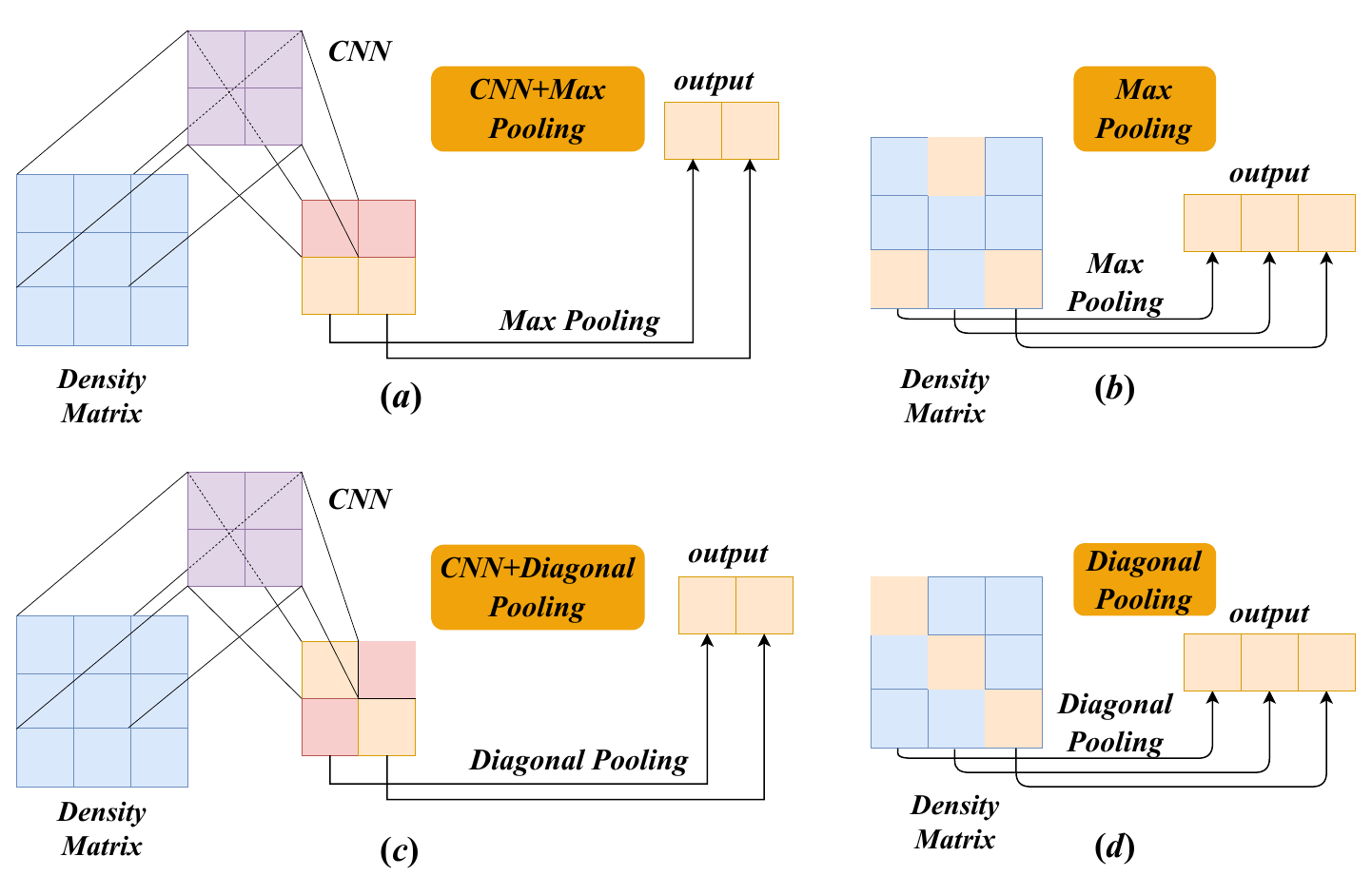}
\caption{Schematic diagram of different feature concentration and extraction operations.}
\label{fig:fig_7}
\end{figure}

The CNN-Diagonal model ranked first with an average score of 1.52, demonstrating superior text classification accuracy by effectively capturing local features and long-distance dependencies. The CNN-Max Pooling model followed closely, ranking second with an average of 1.89. Models using only Max Pooling or Diagonal pooling ranked lower, with averages of 3.37 and 2.05, respectively, though the Diagonal model outperformed Max Pooling in some tasks. These results highlight the critical role of CNNs and the benefits of diagonal pooling in text classification.
\section{Conclusion}
This study explores theoretical and empirical research in text sentiment analysis using quantum-inspired deep learning models. Firstly, a Quantum-Inspired Text Information Representation method is proposed, which efficiently captures semantic features of text using quantum superposition principles and density matrices. Words and sentences are viewed as quantum particles and mixed quantum systems, enhancing the model's interpretability. Secondly, a feature extraction and classification layer that combines long short-term memory with self-attention mechanisms to enhance the model's sentiment judgment capability is designed. It deeply integrates semantic and sentiment information through complex word embedding layers and text density matrix representations. Future research directions include enriching datasets, optimizing the model, expanding the types of embedding data, extending application areas, and enhancing the interpretability of deep learning models, with the expectation of advancing the field.

\FloatBarrier
\section*{Appendices}
\FloatBarrier

In the supplementary material, we first add some description of the chosen dataset. After that some results of the visualizations mentioned in the main text are shown, followed by more graphs of the vector visualization results. After that, we show iterative plots of the evaluation metrics during the training process.

\subsection*{A. Description of the Chosen Dataset}
We describe the categories and content of the selected datasets in detail.
The datasets fall into two categories: those used to evaluate the model's ability to predict the sentiment of sentences, including the Movie Review dataset (MR), the Stanford Sentiment Treebank dataset (SST), the Customer Review dataset (CR), and the Opinion polarity dataset (MPQA); and those used to classify subjective and objective sentences, such as the Subjectivity dataset (SUBJ).

\subsubsection*{Movie Review Dataset (MR).}
The Movie Review dataset (MR) is commonly used for evaluating sentiment analysis models. This dataset consists of movie reviews that have been labeled with their respective sentiment, either positive or negative. The goal is to predict the sentiment of each review based on its textual content.

\subsubsection*{Stanford Sentiment Treebank (SST).}
The Stanford Sentiment Treebank (SST) is a more granular sentiment analysis dataset developed by the Stanford NLP Group. It includes a large number of movie reviews, each labeled not only with an overall sentiment score but also with sentiment annotations at the phrase level. This fine-grained labeling allows models to learn sentiment patterns at different levels of granularity, from single words to entire sentences.

\subsubsection*{Customer Review Dataset (CR).}
The Customer Review dataset (CR) focuses on sentiment analysis in the context of product reviews. It comprises customer reviews from a variety of product categories, each labeled with a sentiment score indicating whether the review is positive or negative. This dataset helps in assessing the performance of sentiment analysis models in understanding and interpreting opinions expressed by customers in their reviews, which is crucial for applications like recommendation systems and market analysis.

\subsubsection*{Opinion Polarity Dataset (MPQA).}
The Opinion Polarity Dataset (MPQA) is designed to evaluate sentiment analysis and opinion mining models. It contains a collection of news articles annotated for opinions and other private states such as beliefs, speculations, and sentiments. The dataset is annotated at multiple levels, including the polarity (positive, negative, or neutral) of expressions, making it useful for fine-grained sentiment analysis tasks.

\subsubsection*{Subjectivity Dataset (SUBJ).}
The Subjectivity Dataset (SUBJ) is used to classify sentences as either subjective or objective. Subjective sentences express personal opinions, emotions, or judgments, while objective sentences present factual information. This dataset contains labeled sentences from various sources, allowing models to learn the characteristics that distinguish subjective content from objective statements. The ability to differentiate between these two types of sentences is important for applications like information retrieval and text summarization, where the nature of the content needs to be accurately identified.
\subsection*{B. Visualization of Sentiment Information Extracted by Wordnet}
\begin{figure*}[ht]
\centering
\includegraphics[width=0.85\linewidth]{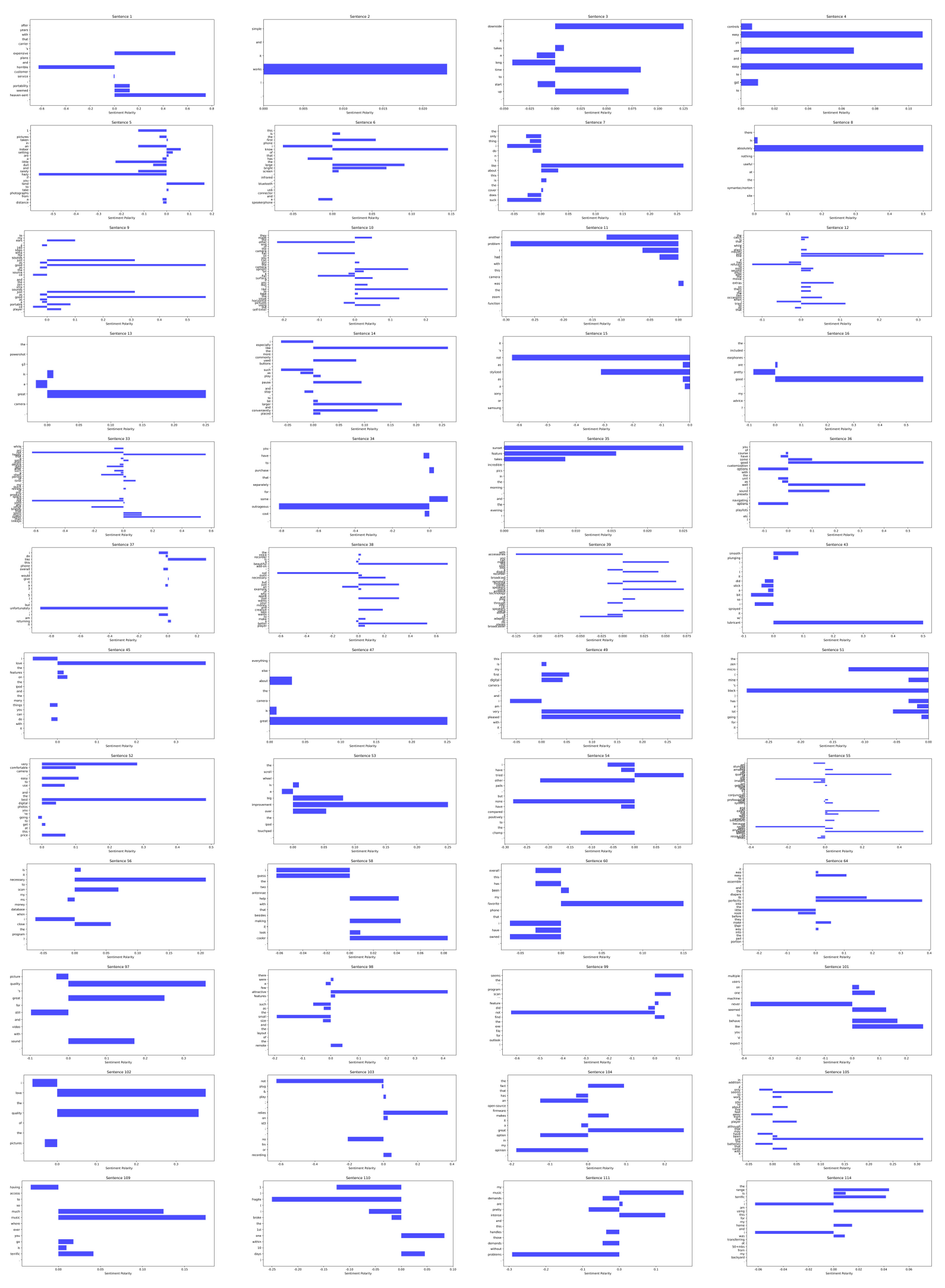}
\caption{\textbf{Sentiment Information of A Sentence Extracted from WordNet.} Using WordNet, we identify the synonym set of a given word and then use SentiWordNet to calculate the sentiment score of each word in the set. By averaging these scores, we derive the overall sentiment score for the given word.}
\label{fig:app_fig_wordnet}
\end{figure*}

We supplemented our analysis with a visualization of the sentiment information extracted using SentiWordNet to verify the validity of the input data, as shown in Fig. \ref{fig:app_fig_wordnet}. The vertical axis represents each word in the sentence, while the horizontal axis indicates the sentiment polarity, which is directly entered as the content of the Phase item without further processing.

\subsection*{C. Visualization of Text Semantic Vectors}
\begin{figure*}[ht]
\centering
\includegraphics[width=0.85\linewidth]{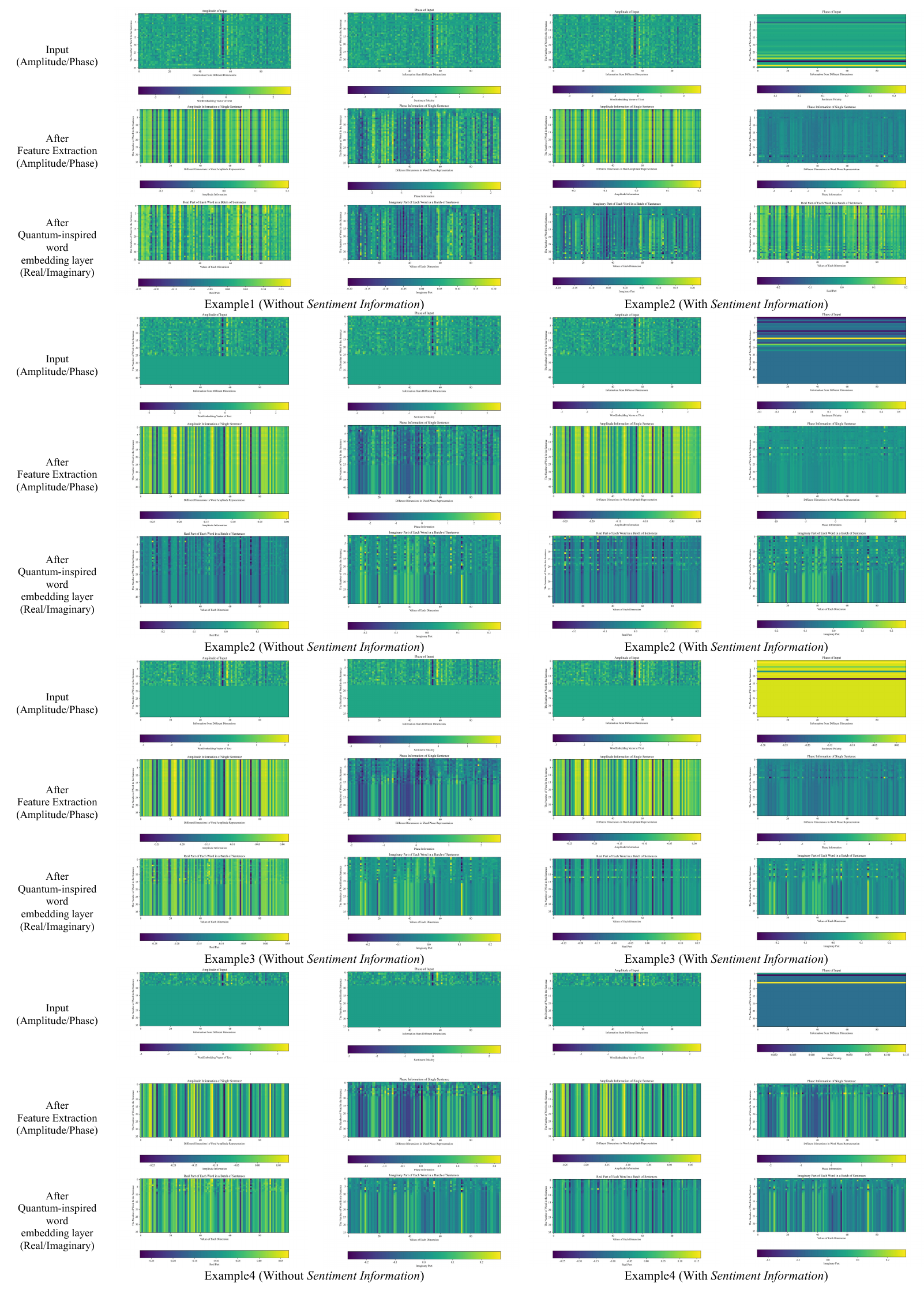}
\caption{\textbf{Additional semantic vector visualization Using SST Dataset.} This image displays two visualization examples from the SST dataset. Notably, semantic vectors enriched with sentiment information exhibit more pronounced changes in the resulting vectors.}
\label{fig:app_fig_SST}
\end{figure*}

\begin{figure*}[hb]
\centering
\includegraphics[width=0.85\linewidth]{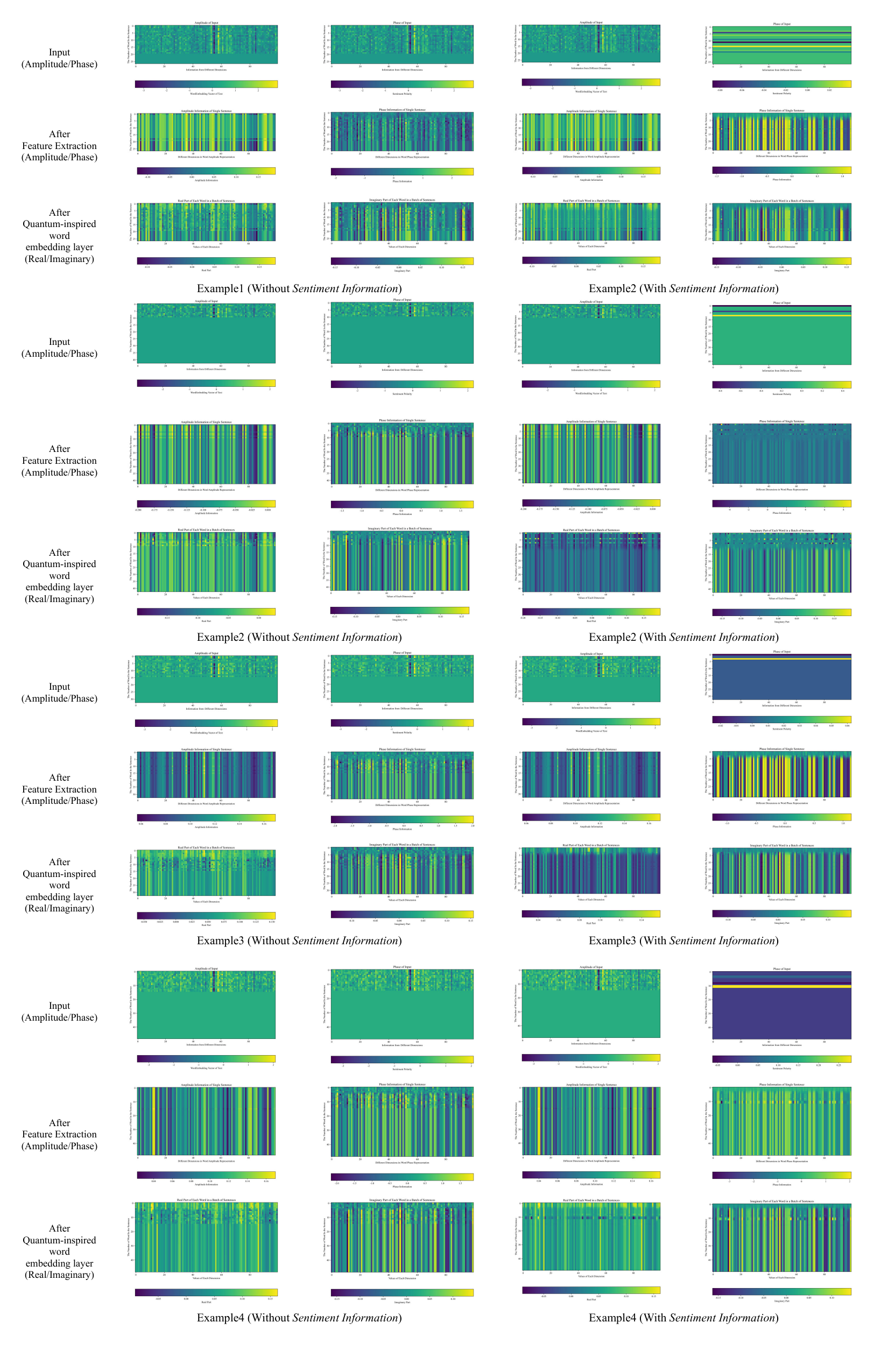}
\caption{\textbf{Additional semantic vector visualization Using CR Dataset.} This image displays two visualization examples from sthe CR dataset. Notably, semantic vectors enriched with sentiment information exhibit more pronounced changes in the resulting vectors.}
\label{fig:app_fig_CR}
\end{figure*}

\begin{figure*}[hb]
\centering
\includegraphics[width=0.85\linewidth]{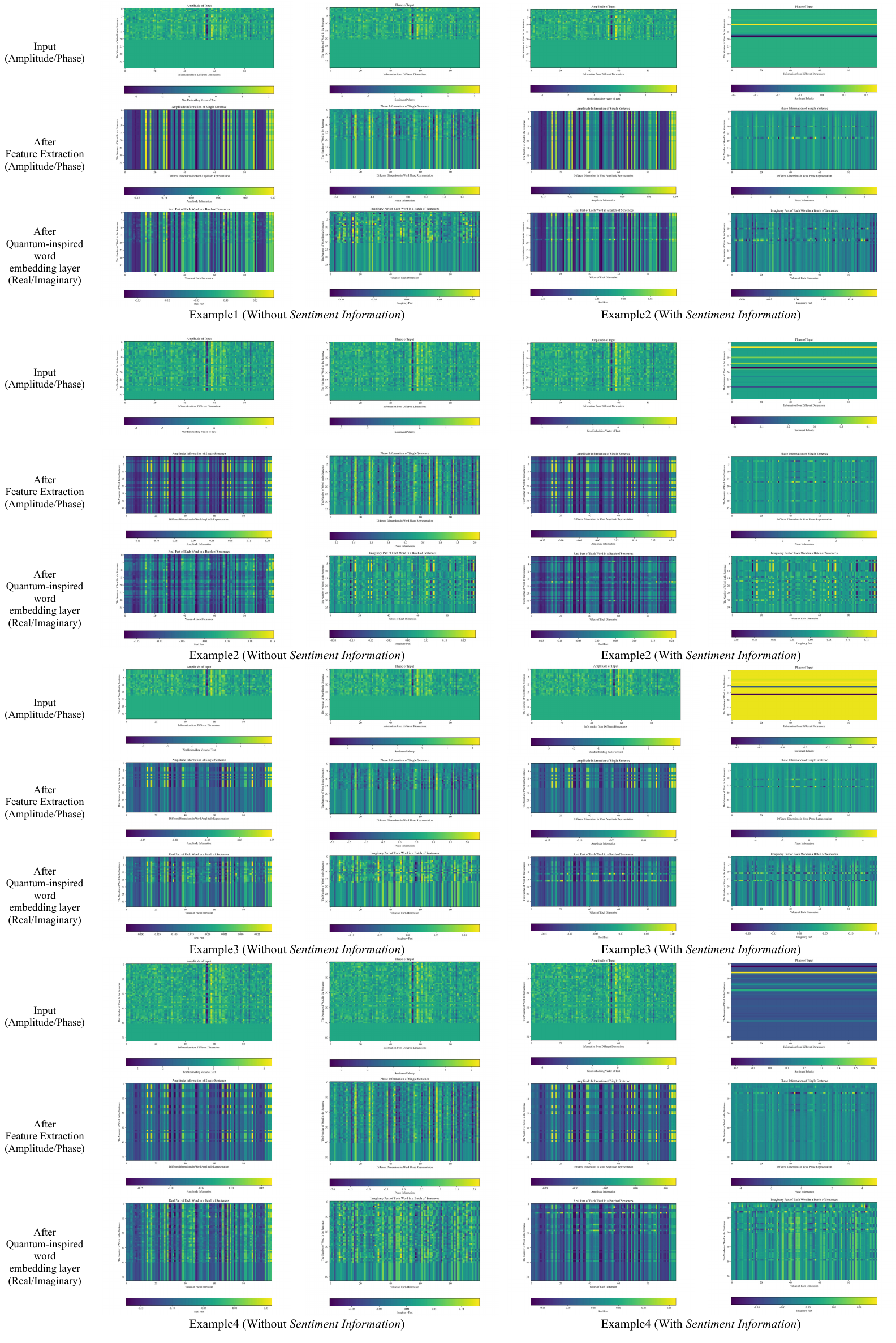}
\caption{\textbf{Additional semantic vector visualization Using MR Dataset.} This image displays two visualization examples from sthe CR dataset. Notably, semantic vectors enriched with sentiment information exhibit more pronounced changes in the resulting vectors.}
\label{fig:app_fig_MR}
\end{figure*}

\begin{figure*}[hb]
\centering
\includegraphics[width=0.82\linewidth]{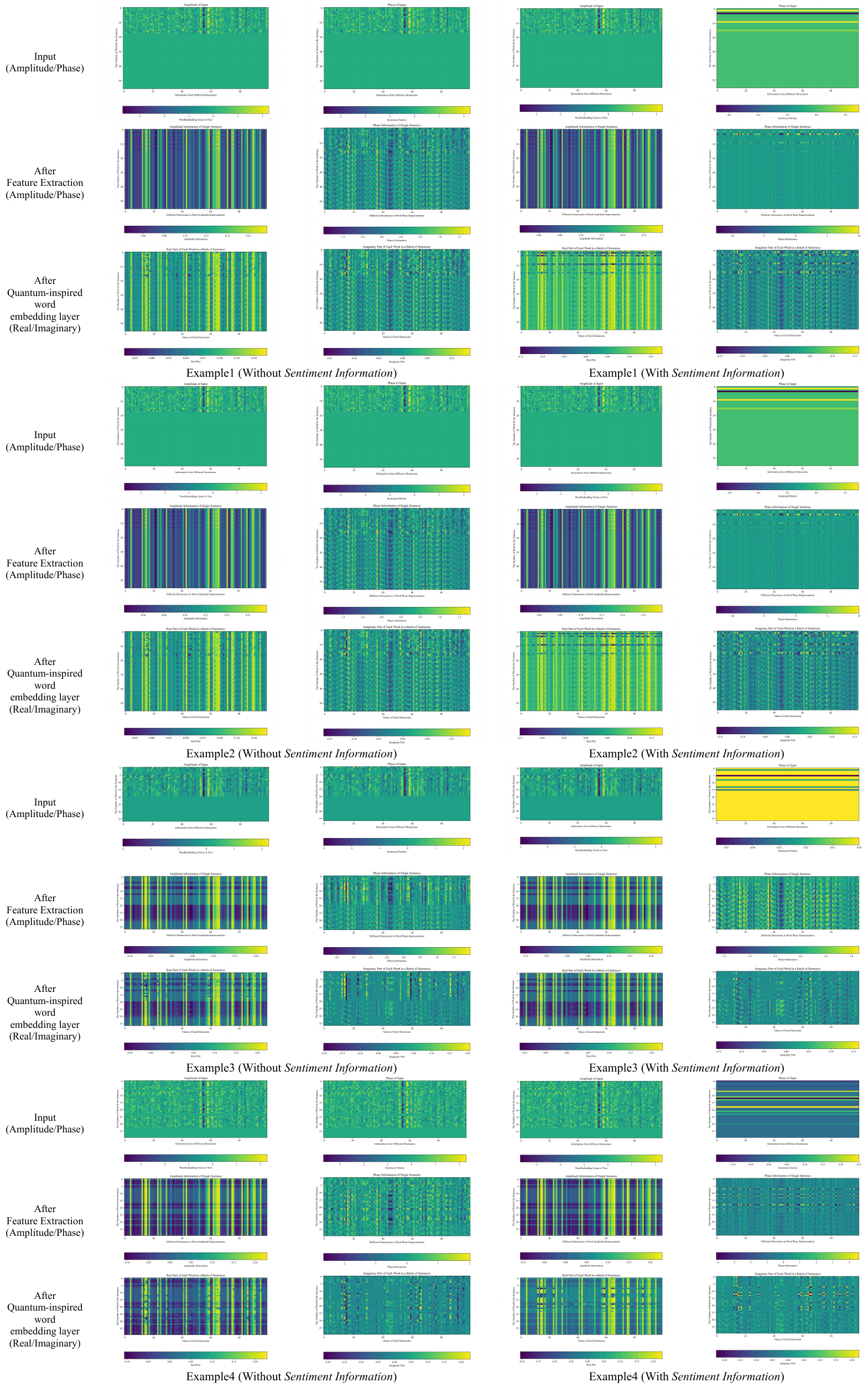}
\caption{\textbf{Additional semantic vector visualization Using SUBJ Dataset.} This image displays two visualization examples from sthe CR dataset. Notably, semantic vectors enriched with sentiment information exhibit more pronounced changes in the resulting vectors.}
\label{fig:app_fig_SUBJ}
\end{figure*}

We conducted further visualization experiments on the aforementioned four datasets, and the experimental results are shown in Fig. \ref{fig:app_fig_SST}--Fig. \ref{fig:app_fig_SUBJ}.
We continue to showcase the visualization of text information vectors in three stages: different input combinations, vectors after feature extraction, and vectors after the quantum-inspired word embedding layer.

In the figures, the horizontal axis represents different inputs of the same sentence, and the vertical axis illustrates various results of vector-matrix visualization. Notably, vectors enriched with sentiment information tend to focus more on words exhibiting significant sentiment polarity changes.
\subsection*{D. Training Curve}
\begin{figure*}[h]
\centering
\includegraphics[width=\linewidth]{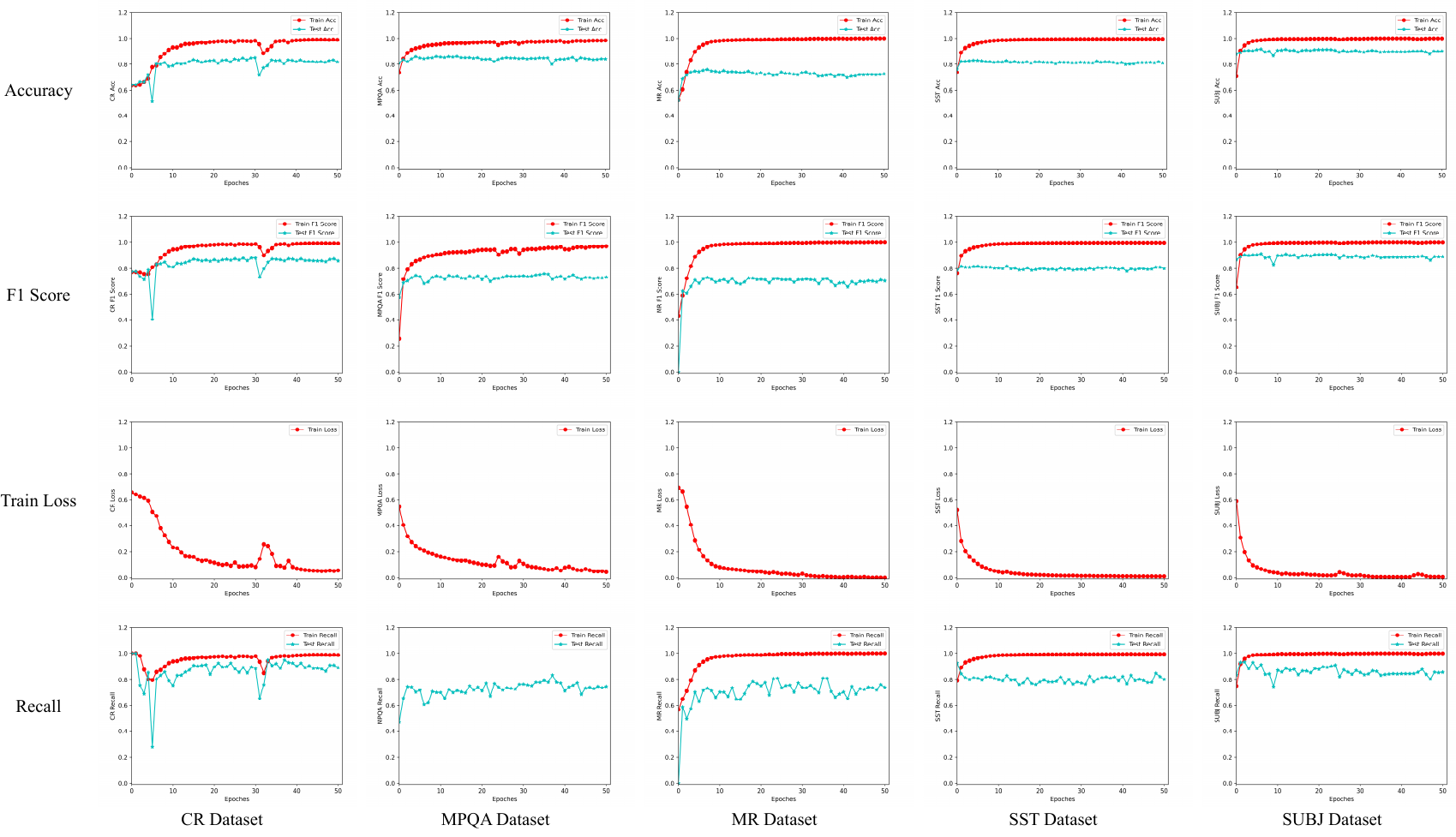}
\caption{\textbf{The iterative process curves of various metrics during training,} where the vertical axis represents different metrics and the horizontal axis denotes different datasets.}
\label{fig:app_fig_2}
\end{figure*}

This section presents train convergence plots of the proposed model on CR, MPQA, MR, SST, and SUBJ datasets, depicted in Fig. \ref{fig:app_fig_2}. 

\subsubsection{Training Loss.} The training loss on 5 datasets, shows a gradual decrease with increasing training epochs, stabilizing in later stages, indicating effective error reduction by the model. However, fluctuations in training loss during mid-training phases, particularly noticeable in CR and MPQA datasets, suggest challenges in handling complex samples.

\subsubsection{Training and test accuracies.} Training and test accuracies show steady improvement and stabilization during training, highlighting the model's learning capability. However, test accuracies generally lag behind training, with noticeable fluctuations in datasets like CR and MR, indicating potential overfitting.

\subsubsection{F1 scores.} F1 scores demonstrate rapid improvement and sustained high performance on training data across all datasets, reflecting balanced precision and recall. Lower test F1 scores and fluctuations in datasets like SUBJ and MPQA suggest challenges in generalization.

\subsubsection{Recall.} Training recall rates improve significantly across all datasets, stabilizing in later stages, demonstrating the model's strength in identifying positive instances. Test recall rates show slower improvement, with fluctuations in datasets like MPQA and MR, indicating varying generalization capabilities.

\FloatBarrier
\FloatBarrier

\end{document}